\documentclass[letterpaper]{article} 
\usepackage{aaai25}  
\usepackage{times}  
\usepackage{helvet}  
\usepackage{courier}  
\usepackage[hyphens]{url}  
\usepackage{graphicx} 
\usepackage{natbib}  
\usepackage{caption} 
\usepackage{algorithm}
\usepackage{algorithmic}
\usepackage{adjustbox}
\usepackage{amsmath}
\usepackage{amsfonts}
\usepackage{amssymb}
\usepackage{soul}
\usepackage{newfloat}
\usepackage{listings}
\usepackage{multirow}
\usepackage{enumitem}
\usepackage{threeparttable}
\usepackage{url}

\usepackage{newfloat}
\usepackage{listings}
\usepackage{multirow}
\usepackage{diagbox}
\usepackage{graphicx}
\usepackage{booktabs}
\usepackage{enumitem}
\usepackage{color}
\usepackage{multirow}
\usepackage{array}
\usepackage{amsmath}
\usepackage{pifont}

\usepackage{multirow}
\usepackage{diagbox}
\usepackage{graphicx}
\usepackage{booktabs}
\usepackage{enumitem}
\usepackage{color}
\usepackage{multirow}
\usepackage{array}
\usepackage{amsmath}
\usepackage{pifont}
\usepackage{xcolor}
\usepackage{graphicx}
\usepackage{subcaption}

\DeclareCaptionStyle{ruled}{labelfont=normalfont,labelsep=colon,strut=off} 
\floatstyle{ruled}
\newfloat{listing}{tb}{lst}{}
\floatname{listing}{Listing}
\title{Unlocking the Power of LSTM for Long Term Time Series Forecasting}

\author{
    Yaxuan~Kong\textsuperscript{\rm 1}\equalcontrib,
    Zepu~Wang\textsuperscript{\rm 2 }\textsuperscript{\rm 5}\equalcontrib,
    Yuqi~Nie\textsuperscript{\rm 3},
    Tian~Zhou\textsuperscript{\rm 4},
    Stefan~Zohren\textsuperscript{\rm 1},
    Yuxuan~Liang\textsuperscript{\rm 2}\thanks{Corresponding authors.},
    Peng~Sun\textsuperscript{\rm 5}$^\dagger$,
    Qingsong~Wen\textsuperscript{\rm 6}
}
\affiliations{
    \textsuperscript{\rm 1}University of Oxford \\
    \textsuperscript{\rm 2}The Hong Kong University of Science and Technology (Guangzhou) \\
    \textsuperscript{\rm 3}Princeton University \\
    \textsuperscript{\rm 4}Alibaba Group \\
    \textsuperscript{\rm 5}Duke Kunshan University \\
    \textsuperscript{\rm 6}Squirrel AI \\
    yaxuan.kong@eng.ox.ac.uk,
    zw163@duke.edu
}

\usepackage{bibentry}

\begin{document}

\maketitle

\begin{abstract}
Traditional recurrent neural network architectures, such as long short-term memory neural networks (LSTM), have historically held a prominent role in time series forecasting (TSF) tasks. While the recently introduced sLSTM for Natural Language Processing (NLP) introduces exponential gating and memory mixing that are beneficial for long term sequential learning, its potential short memory issue is a barrier to applying sLSTM directly in TSF. To address this, we propose a simple yet efficient algorithm named P-sLSTM, which is built upon sLSTM by incorporating patching and channel independence. These modifications substantially enhance sLSTM's performance in TSF, achieving state-of-the-art results. Furthermore, we provide theoretical justifications for our design, and conduct extensive comparative and analytical experiments to fully validate the efficiency and superior performance of our model.
\end{abstract}

%
\begin{links}
    \link{Code}{https://github.com/Eleanorkong/P-sLSTM}
\end{links}

\section{Introduction} 

Time Series Forecasting (TSF) is a prominent area of research in statistics and machine learning with diverse applications ranging from financial forecasting to traffic and human trajectory prediction~\cite{wang2022novel, wang2023st, wang2022novel1, nie2024survey, zhao2024hedge, guo2024hybrid, ruan2024infostgcan}. Historically, traditional Recurrent Neural Networks (RNNs) have been the cornerstone for modeling sequential data due to their inherent ability to capture temporal dependencies. Among different RNN architectures, Long Short-Term Memory (LSTM) networks are among the most popular and successful ones, as they solve the traditional vanishing and exploding gradient problems of RNNs and exhibit better performance with long sequences. Numerous researchers have adopted LSTM networks to predict diverse time series data, demonstrating their effectiveness across various domains~\cite{hochreiter1997long,gers1999learning,siami2018comparison}.

However, LSTM has several limitations in TSF. Firstly, although LSTM is designed to capture longer sequential correlations compared to traditional RNNs, it fails to memorize long sequential information statistically and experimentally~\cite{zhao2020rnn}. Additionally, LSTMs lack the ability to revise storage decisions dynamically with their historical forget gates, which restricts their flexibility to adapt to changes in data~\cite{oreshkin2019n,beck2024xlstm,zhao2020rnn}. Consequently, traditional RNNs have gradually lost their dominant position in time series tasks, with the leaderboard for TSF now being dominated by alternative architectures such as Transformers~\cite{wen2023transformers,zhou2021informer} and Multi-Layer Perceptrons (MLPs)~\cite{wang2023st,zeng2023transformers}.

\begin{table}[ht]
\centering
\scriptsize 
\begin{tabular}{lccc}
\hline
Method & Type & Time Complexity  & Memory Complexity  \\
\hline
LSTM & RNN & $O(L)$ & $O(L)$ \\
DLinear & MLP  &$O(L)$ & $O(L)$  \\

Informer & Transformer &$O(L\log L)$ & $O(L\log L)$  \\
Crossformer & Transformer & $O(L^2)$ & $O(L^2)$  \\
\hline
\end{tabular}
\caption{Time and Space Complexity Comparison of TSF Models. $L$ is the input sequence length/look-back window. LSTM~\cite{hochreiter1997long}; DLinear~\cite{zeng2023transformers}; Informer~\cite{zhou2021informer}; Crossformer~\cite{zhang2023crossformer}.}

\label{Table 1: Time Complexity}
\end{table}

Despite these limitations, LSTM and, more broadly, RNNs remain powerful tools and a promising research direction due to their advantages~\cite{hewamalage2021recurrent}. Firstly, compared to most transformer-based models, RNNs have much lower time complexity and memory complexity, making them more efficient for certain applications, as shown in Table~\ref{Table 1: Time Complexity}. Secondly, RNNs have a clear temporal flow, which makes it easier to interpret their decisions and understand how information flows through the sequence compared to transformer-based structures and MLP-based structures~\cite{graves2013generating,hou2020learning}.
At the same time, state-space models (SSMs) have received wide attention in TSF and broader deep learning research areas. Compared to other black-box deep learning models, SSMs can be explained from statistical and physical perspectives, potentially offering better interpretability~\cite{gu2022parameterization,wang2024mamba}. Researchers have shown that LSTM/RNNs can be viewed as a special case of SSMs, where the recurrent process illustrates the flow of information~\cite{kim1994nonlinear,durbin2012time,zhao2020rnn}. Therefore, LSTM/RNNs are also a reasonable research direction, through the lens of those findings.


Recently, an advanced version of the extended LSTM, named sLSTM~\cite{beck2024xlstm}, was introduced showing that one can not only revise the memory storage decisions but also improve its memory capacities, leading to very competitive performance in various NLP tasks~\cite{beck2024xlstm}. Given the success of advanced LSTMs in NLP, can we unlock the power of LSTMs for time series forecasting? In this paper, we give an affirmative answer by reproposing sLSTM for multivariate TSF, leading to our new method called P-sLSTM. Our main contributions are as follows.

\begin{itemize}
\item[$\bullet$] We explain why the framework of sLSTM improves memory capacity and revises memory storage, which is suitable for TSF;
\item[$\bullet$] We rigorously show that sLSTM cannot be guaranteed to have a long memory to capture long-term dependencies. Based on the previous limitation, we apply the patching technique to solve this problem and develop an LSTM-based structure, P-sLSTM, for TSF; 
\item[$\bullet$] Extensive evaluations on various datasets shows that P-sLSTM outperformaces the original LSTM by 20\% of accuracy and reaches comparable performance against state-of-the-art (SOTA) models.

\end{itemize}

\section{Related Work}
\label{Ch: Literature Review}

TSF has seen significant advancements with the introduction of various model architectures such as Transformer~\cite{lin2023comprehensive}, and MLP~\cite{lu2024cats}. Transformer-based models have gained popularity in TSF due to their ability to capture long-range dependencies and handle complex temporal patterns~\cite{wen2023transformers,lim2021temporal}. For instance, the Informer model by ~\citet{zhou2021informer} improves the efficiency and scalability of Transformers by using a sparse self-attention mechanism and a generative style decoder, allowing it to handle long sequence forecasting more effectively. Additionally, many other Transformer-based models are developed for TSF, such as Autoformer~\cite{wu2021autoformer}, Fedformer~\cite{zhou2022fedformer}, iTransformer~\cite{liu2023itransformer}. 


In comparison, MLP-based models typically involve feeding lagged time series data into a fully connected neural network to predict future values ~\cite{zeng2023transformers}. For instance, DLinear combines decomposition techniques to get the seasonal information and trend information of the time series data and uses MLP layers to get a good prediction accuracy~\cite{zeng2023transformers}. Another example is Rlinear, which applies linear mapping to the data before linear layers~\cite{li2023revisiting}.


State space models (SSMs) have gained prominence due to their inherent flexibility and interpretability in TSF~\cite{Durbin2012}. SSMs capture temporal dynamics by representing time series as a combination of observable and hidden states that evolve over time through specified transition equations~\cite{gu2022parameterization,wang2024mamba,hu2024time}.
One notable framework leveraging SSMs is the Mamba model, which employs a hierarchical Bayesian approach to capture intricate temporal patterns and adapt to changing dynamics over time~\cite{gu2023mamba}.

\section{sLSTM for Time Series Forecasting}

In this section, we define the TSF problem and revisit the general concept of LSTM. We then briefly explain why the new sLSTM framework can revise historical memory storage and why it is suitable for time series data.

\subsection{Problem Formulation}
\label{Ch: Problem Statment}

The objective of multivariate TSF is to learn a mapping function \( f \) that takes historical data \( X \) with \( M \) channels as input, denoted as \( X \in \mathbb{R}^{L \times M} \), where \( L \) represents the look-back window. The goal is to predict future data for a time interval of length \( T \), denoted as \( \hat{X} \in \mathbb{R}^{T \times M} \).

\subsection{Recap of LSTM Structure}
The LSTM \cite{hochreiter1997long, hochreiter2001gradient} was original introduced to avoid the vanishing gradient problem encountered in traditional RNNs. The LSTM memory cell contains three gates: input, output, and forget gates. The LSTM memory cell update rules at time step \( t \) are as follows:

{\footnotesize
\begin{align}
    \mathbf{c}_t &= \mathbf{f}_t \odot \mathbf{c}_{t-1} + \mathbf{i}_t \odot \mathbf{z}_t \quad \text{cell state} \label{eq:cell_state}\\
    \mathbf{h}_t &= \mathbf{o}_t \odot \tilde{\mathbf{h}}_t, \quad \tilde{\mathbf{h}}_t = \psi(\mathbf{c}_t) \quad \text{hidden state} \label{eq:hidden_state}\\
    \mathbf{z}_t &= \varphi(\tilde{\mathbf{z}}_t), \quad \tilde{\mathbf{z}}_t = \mathbf{W}_z \mathbf{x}_t + \mathbf{R}_z \mathbf{h}_{t-1} + \mathbf{b}_z \quad \text{cell input} \label{eq:cell_input}\\
    \mathbf{i}_t &= \sigma_{lstm}(\tilde{\mathbf{i}}_t), \quad \tilde{\mathbf{i}}_t = \mathbf{W}_i \mathbf{x}_t + \mathbf{R}_i \mathbf{h}_{t-1} + \mathbf{b}_i \quad \text{input gate} \label{eq:input_gate}\\
    \mathbf{f}_t &= \sigma_{lstm}(\tilde{\mathbf{f}}_t), \,\,\tilde{\mathbf{f}}_t = \mathbf{W}_f \mathbf{x}_t + \mathbf{R}_f \mathbf{h}_{t-1} + \mathbf{b}_f \,\, \text{forget gate} \label{eq:forget_gate}\\
    \mathbf{o}_t &= \sigma_{lstm}(\tilde{\mathbf{o}}_t), \,\,\tilde{\mathbf{o}}_t = \mathbf{W}_o \mathbf{x}_t + \mathbf{R}_o \mathbf{h}_{t-1} + \mathbf{b}_o \,\, \text{output gate} \label{eq:output_gate}
\end{align} 
} 

In this structure, the input weight matrices ($\mathbf{W}_z, \mathbf{W}_i, \mathbf{W}_f,$ and $\mathbf{W}_o$) link the input $\mathbf{x}_t$ to the cell input and the three gates (input, forget, and output). The recurrent weight matrices ($\mathbf{R}_z, \mathbf{R}_i, \mathbf{R}_f,$ and $\mathbf{R}_o$) connect the previous hidden state $\mathbf{h}_{t-1}$ to the cell input and the same three gates. Each component also has its associated bias vector ($\mathbf{b}_z, \mathbf{b}_i, \mathbf{b}_f,$ and $\mathbf{b}_o$). The functions $\varphi$ and $\psi$ are activation functions for the cell input and hidden state, respectively. Typically, both use the hyperbolic tangent ($\tanh$) function. The purpose of $\psi$ is to constrain the cell state within a specific range, preventing it from growing without bounds. In the standard LSTM network, the activation function for all gates is the sigmoid function.

\subsection{sLSTM}
The sLSTM (scalar LSTM) has recently been introduced as part of the xLSTM (extend LSTM) family~\cite{beck2024xlstm}. It is designed to enhance the LSTM architecture by incorporating exponential gating and a new memory mixing mechanism. Compared to the traditional LSTM architecture, the sLSTM introduces several key differences:

\begin{itemize}
    \item[$\bullet$] Forget Gate: The sLSTM allows for either sigmoid or exponential activation function in the forget gate, whereas traditional LSTM uses sigmoid. This is represented as:
    \begin{equation}
        \mathbf{f}_t =  \exp(\tilde{\mathbf{f}}_t)
    \end{equation}

    \item[$\bullet$] Input Gate: The sLSTM uses an exponential function for the input gate, replacing the sigmoid function used in traditional LSTM units:
    \begin{equation}
        \mathbf{i}_t = \exp(\tilde{\mathbf{i}}_t)
    \end{equation}

    \item[$\bullet$] Normalizer State and Hidden State: The sLSTM introduces a normalizer state $\mathbf{n}_t$ and modifies the hidden state calculation:
    \begin{equation}
        \mathbf{h}_t = \mathbf{o}_t \odot \tilde{\mathbf{h}}_t, \quad \tilde{\mathbf{h}}_t = \mathbf{c}_t \odot {\mathbf{n}_t}^{-1}
    \end{equation}
    where the normalizer state is updated as:
    \begin{equation}
        \mathbf{n}_t = \mathbf{f}_t \odot \mathbf{n}_{t-1} + \mathbf{i}_t
    \end{equation}
\end{itemize}

These modifications on the exponential gating in both the forget and input gates provides more flexibility in controlling information flow, while the normalizer state helps in stabilizing the hidden state calculations over long sequences.

The sLSTM architecture further introduces a multi-head structure and a new memory mixing mechanism. Each sLSTM block can have multiple heads, with memory mixing occurring within each head but not across heads. This is achieved through block-diagonal recurrent weight matrices, allowing for efficient parameter sharing while maintaining separate information flows. The memory mixing process can be represented as $\mathbf{h}_t = f(\mathbf{x}_t, \mathbf{h}_{t-1}, \mathbf{c}_{t-1}, \mathbf{n}_{t-1})$, where $f$ encapsulates all sLSTM operations including the block-diagonal connections and exponential gating. Multiple sLSTM blocks can be stacked together to form deeper networks, enhancing the model's capacity to capture complex temporal dependencies in sequential data (see sLSTM structure in Figure \ref{Figure 2: Overall Structure}).

\subsection{Why Exponential Gating?}
The introduction of exponential gating in sLSTM enables the model to revise memory storage decisions compared to the standard LSTM. Unlike the sigmoid activation function, exponential gating allows for a broader range of values, providing the forget gate with more diverse outputs in different situations. This is beneficial in TSF, where the importance of past information can change dynamically, and the model could revise stored values when encountering more relevant inputs. Moreover, exponential gating helps mitigate the vanishing gradient problem often encountered in sigmoid-based gates. Since the derivative of the exponential function is the function itself, gradients can flow more easily during backpropagation, even in long sequences. 


\subsection{Why Memory Mixing?}
The sLSTM model also introduces a memory mixing mechanism to incorporate multiple heads, which are diagonal blocks within the recurrent gate pre-activations, and exponential gating. Both of them establish a new approach to memory mixing. This feature enables the model to dynamically combine and integrate memories from different time steps, allowing it to retain and utilize important information over extended periods more effectively. The relevance of historical data can vary, and memory mixing provides an advantage by dynamically adjusting the importance of past information. This flexibility allows the model to adapt to changing patterns and trends in the data and capture complex temporal dependencies. Memory mixing ensures that relevant historical data, regardless of its temporal distance, can be effectively integrated into the forecasting process. This leads to more informed predictions that take into account both short-term fluctuations and long-term trends.


\section{Methodology}
In this section, we will begin by providing a theoretical foundation for memory property of sLSTM. Next, we will define sLSTM Markov Chain process and provide detailed theoretical analysis on memory property of sLSTM. Then, we design a simple yet efficient algorithm named P-sLSTM, built upon sLSTM with patching and channel independence. 





\subsection{Theoretical Preparation}

Let us first recall some basic property results related to RNN/LSTM~\cite{zhao2020rnn}.
\\~\\ 
\textbf{Assumption 1.}
\textit{(i) The joint density function of $\epsilon^{(t)}$ is continuous and positive everywhere. Note that $\epsilon^{(t)}$ is a sequence of independent and identically distributed random vectors, where the target sequence $y^{(t)}$ $=$ the output term $e^{(t)}$ $+$ the error term $\epsilon^{(t)}$. (ii) For some $\kappa \ge 2$, $\mathbb{E}\|\epsilon^{(t)}\|^\kappa < \infty$. }

\vspace{0.5em} 
\noindent\textbf{Definition 1.}
\textit{A process \(\{X_t\}\) is geometrically ergodic if it converges exponentially fast over time. From Harris Theorem \cite{meyn2012markov}, \(\gamma_X(k) \sim \rho^k\) as \(k \to \infty\), indicating short memory. This means the influence of past observations on future distributions diminishes exponentially fast.}

\vspace{0.5em} 
\noindent\textbf{Theorem 1.} 
\textit{Under Assumptions 1, if there exist real numbers $0 < a < 1$ and $b$ such that $\| \mathcal{M}(x) \| \leq a \| x \| + b$, then recurrent network process with transition function $\mathcal{M}(x)$ is geometrically ergodic, and hence has short memory.}\footnote{We formally define the transition function $\mathcal{M}(x)$ in subsection \textbf{sLSTM Process}.}
\begin{figure*}[ht]
\begin{center}
\includegraphics[width = \linewidth]{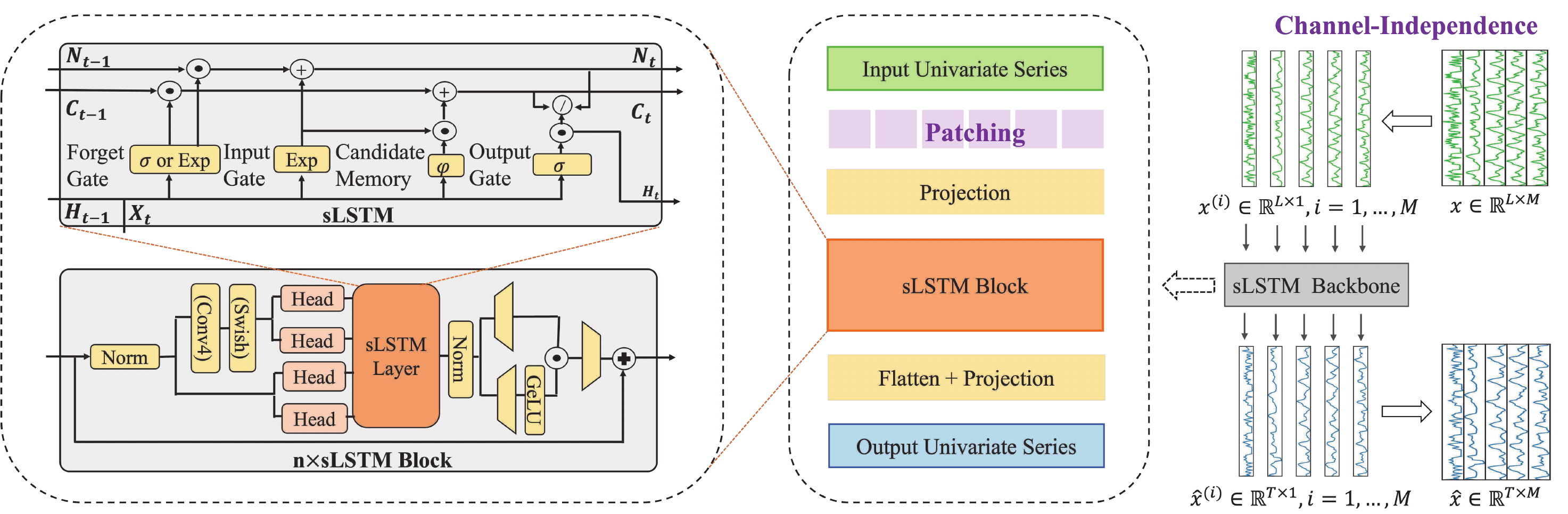}
\end{center}
\caption{Overview of P-sLSTM Architecture (Top Left: sLSTM structure; Bottom Left: sLSTM block~\cite{beck2024xlstm}).}
\label{Figure 2: Overall Structure}
\end{figure*}

\subsection{sLSTM Process}
Many time series models, including LSTM and general RNNs, can be viewed as state-space models and transformed into Markov Chain processes. \citet{zhao2020rnn} reformulated the LSTM as a Markov Chain process and proved, under mild conditions, that it exhibits geometric ergodicity and short-term memory characteristics. To examine the memory properties of sLSTM, we will define an sLSTM Markov Chain process. We adopt a framework similar to that outlined by \cite{zhao2020rnn} for this definition.

Consider a many-to-many sLSTM structure for TSF problems, using square loss as an example:

\begin{equation}
\begin{cases}
    l^{(t)} = \|y^{(t)} - e^{(t)}\|^2 & \text{loss} \\
    e^{(t)} = g(W_{eh} h^{(t)} + b_e) & \text{output} \\
    h^{(t)} = o^{(t)} \odot \left( \frac{c^{(t)}}{n^{(t)}} \right) & \text{hidden state}
\end{cases}
\end{equation}
where the hidden unit can be calculated using the sLSTM gating operations. Note here that as mentioned in \citet{zhao2020rnn}, including long-range dependencies in an external input could interfere with our following analysis of the memory characteristics of $\{y^{(t)}\}$. Thus, for simplicity, we also assume no external input, meaning $x^{(t)} = y^{(t-1)}$. The process is defined for $t \in \{1, ..., T\}$, where $y^{(0)} = 0$, $h^{(0)} = 0$, $y^{(t)}, e^{(t)} \in \mathbb{R}^p$, $h^{(t)}, c^{(t)}, z^{(t)}, f^{(t)}, i^{(t)}, o^{(t)} \in \mathbb{R}^q$, $g$ is an elementwise output function, and $\odot$ denotes elementwise product.


For sLSTM process, we define the hidden states $s^{(t)} \in \mathbb{R}^q$ to include the cell state $c^{(t)}$, the normalizer state $n^{(t)}$, and the hidden state $h^{(t)}$. The sLSTM process can then be represented as a homogeneous Markov chain with the transition function $\mathcal{M}_\text{sLSTM}$:

{\footnotesize
\begin{equation}
\begin{pmatrix}
y^{(t)} \\
h^{(t)} \\
c^{(t)} \\
n^{(t)}
\end{pmatrix}
\!\!=\!\!
\mathcal{M}_\text{sLSTM}
\begin{pmatrix}
y^{(t-1)},
h^{(t-1)}, 
c^{(t-1)}, 
n^{(t-1)}
\end{pmatrix}
\!+\!\!
\begin{pmatrix}
\epsilon^{(t)} \\
0 \\
0 \\
0
\end{pmatrix}
\label{eq:sLSTM Markov Chain}
\end{equation}
} 

In the above formulation, the transition function $\mathcal{M}_\text{sLSTM}$ maps $(\mathbb{R}^p \times \mathbb{R}^q \times \mathbb{R}^q \times \mathbb{R}^q)$ to $(\mathbb{R}^p \times \mathbb{R}^q \times \mathbb{R}^q \times \mathbb{R}^q)$, where $q$ is the dimension of the hidden state vectors: $h^{(t)}$, $c^{(t)}$, $n^{(t)}$.

\subsection{Memory Property of sLSTM Process}
\citet{zhao2020rnn} analysis revealed that the forget gate plays a crucial role in controlling the output range of the forget gate and determining the memory properties of LSTM. Notably, the primary distinction between LSTM and sLSTM lies in the activation function of the forget gate: LSTM typically employs a sigmoid function, while sLSTM uses the exponential function. Intuitively, in sLSTM, the range of the forget gate with an exponential activation function is unbounded (please refer to Case 2). Consequently, an exponentially large output from the forget gate suggests a greater retention of past information. This characteristic allows the model to better capture long-term dependencies in time series data, thereby enhancing its memory capacity.

However, does this mean that sLSTM is guaranteed to have a long memory? Building upon \citet{zhao2020rnn}'s work, we extended the analysis from LSTM to sLSTM and introduced an additional corollary based on Assumption 1 and Theorem 1. We prove that under mild conditions, the memory loss property is preserved as the sLSTM process also maintains geometric ergodicity, indicating that it potentially has a short memory. Specifically, this short memory property holds under the condition that \( ||f(u,v)||_{\infty} \leq a \) for some \( a < 1 \). For cases where this condition exceeds 1, we examined the long-term behavior of the forget gate to understand its implications for memory retention.
\subsubsection{Case 1:} When \( ||f(u,v)||_{\infty} \leq a \) for some \( a < 1 \), we observe that the sLSTM network exhibits geometric ergodicity, and hence short memory.
\\~\\
\textbf{Corollary.} \textit{The input series features \(\{y^{(t-1)}\}\) are scaled to the range \([-1, 1]\). Suppose that $M := \sup_{\mathbf{x} \in B^q_\infty} \|g(W_{eh} \mathbf{x} + b_e)\|_{l1} < \infty$
and \(\|f(u, v)\|_{l_\infty} = \|\exp(W_f u + R_f v + b_f)\|_{l_\infty} \le a\) for some \(a < 1\), where \(B^q_\infty\) is the \(q\)-dimensional \(l_\infty\)-ball and \(\|W\|_{l_\infty} = \max_{1 \le i \le m} \sum_{j=1}^n |w_{ij}|\) is the matrix \(l_\infty\)-norm. If Assumption 1 holds, then the sLSTM process is geometrically ergodic and has short memory.}
\\~\\
\textit{Proof:} 
Let \( a = a_0 \in (0, 1) \) and \( b = M + 3q \), then: \\
$$\|\mathcal{M}_{\text{sLSTM}} (u, v, w, n)\|_{l1} - a_0 \| (u', v', w', n')\|_{l1}$$
$ \le \|g(W_{eh} x + b_e)\|_{l1} + \|x\|_{l1} 
+ \| \mathbf{1}_q + f(u, v) \odot w \|_{l1} $
$$+ \|f(u, v) \odot n + \mathbf{1}_q\|_{l1} 
- a_0 \| (u', v', w', n')\|_{l1} $$
$ \le M + q + q + \|f(u, v)\|_{l_\infty}\|{w}\|_{l1}
+ \|f(u, v)\|_{l_\infty}\|n\|_{l1} + q $
$$- a_0\|u\|_{l1} - a_0\|v\|_{l1} 
- a_0\|w\|_{l1} - a_0\|n\|_{l1} $$
$ \le M + 3q - a_0\|u\|_{l1} - a_0\|v\|_{l1}
+ (\|f(u, v)\|_{l_\infty} - a_0)\|w\|_{l1} $
$$ + (\|f(u, v)\|_{l_\infty} - a_0)\|n\|_{l1} 
 \le b = M + 3q, $$
where
$ x = o(u, v) \odot [i(u, v) \odot \tanh(R_{z} v + W_{z} u + b_z) $
$ + f(u, v) \odot w] \odot \left[f(u, v) \odot n + i(u, v)\right]^{-1} \in B^q_\infty\ , $
\(v = h_t \in B^q_\infty\), and \(u = y_{t-1} \in B^p_\infty\). 
The second inequality holds due to the definition of \(M\) and \(x \in B^q_\infty\). The fourth inequality holds due to $\|f(u, v)\|_{l_\infty} = \|\exp(W_f u + R_f v + b_f)\|_{l_\infty} \le a_0. $
By Assumption 1 and Theorem 1 in \citet{zhao2020rnn}, the sLSTM model (\ref{eq:sLSTM Markov Chain}) is geometrically ergodic and has short memory if \(\|f(u, v)\|_{l_\infty} \le a\) for some $a < 1$. 
\hfill $\square$
\\~\\
The above proof aligns with Zhao's (2020) observations on LSTM, where we extend the analysis to sLSTM, showing that the forget gate primarily influences the memory property of sLSTM. As the geometric ergodicity of sLSTM only holds when $||f(u, v)||{l_\infty} = ||\exp(W_f u + R_f v + b_f)||{l_\infty} \le a$ for some $a < 1$, this implies that $W_f u + R_f v + b_f$ need to be negative (since the exponential function of a negative value is less than 1). This condition is analogous to the use of a sigmoid activation function on the forget gate in standard LSTM. 

\subsubsection{Case 2: \label{Case 2}} When \( ||f(u,v)||_{\infty} > 1 \), we examine the long term behavior of the exponential forget gate of sLSTM. 
\\~\\
It is also important to consider the case when $||f(u, v)||{l_\infty}$ exceeds 1. This is when $W_f u + R_f v + b_f$ is positive, leading to an exponential value greater than 1. 

In this scenario, the forget gate $f_t = \exp(\tilde{f}_t)$ tends to amplify the cell state $c_t$ over time instead of allowing it to decay. This behavior seems to revise the memory storage and assign larger values to new information that is important; however, it may diminish the model's ability to effectively integrate and prioritize new information over time.

The cell state update equation $c_t = f_t \odot c_{t-1} + i_t \odot z_t$ shows that when $f_t > 1$ and grows larger, the past state's contribution $(f_t \odot c_{t-1})$ becomes larger than the new input information $(i_t \odot z_t)$. This leads to exponential growth of both $c_t$ and the normalizer state $n_t$ $(n_t = f_t \odot n_{t-1} + i_t)$ over time, potentially resulting in unbounded values.

As $f_t$ becomes larger, the impact of the input gate $i_t \odot z_t$ on the cell state and normalizer state diminishes, as the contribution from the past state dominates. In this case, we can approximate that $c_t \approx f_t \odot c_{t-1}$ and $n_t \approx f_t \odot n_{t-1}$.

Consequently, the model's capacity to learn is diminished. It may struggle to integrate new information effectively as the cell state becomes dominated by amplified past states. Furthermore, the hidden state calculation $h_t = o_t \cdot \frac{c_t}{n_t}$ may face precision problems due to the exponential growth of both $c_t$ and $n_t$. 

Even if $c_t$ and $n_t$ grow at the same rate, and their ratio $\frac{c_t}{n_t}$ $(\frac{c_t}{n_t} \approx \frac{f_t \odot c_{t-1}}{f_t \odot n_{t-1}} = \frac{c_{t-1}}{n_{t-1}})$ might remain stable, the absolute values can become very large, leading to overflow issues.
\\~\\
The above analysis shows that the memory properties of sLSTM are closely tied to the behavior of its forget gate. When the forget gate's output is bounded below 1, sLSTM exhibits geometric ergodicity and short memory, similar to standard LSTM. However, when the forget gate's output exceeds 1, it can lead to exponential growth in the cell state, potentially causing computational issues and diminishing the model's ability to effectively integrate new information.

\subsection{General Discussion of the RNN memory property}
Is it possible to design an LSTM or a general RNN framework that has a long memory? Theoretically, we might be able to. However, in practice, it is quite hard to ensure that we can use neural networks to reach this goal.

Using LSTM as an example, to maintain long-term memory, we need to ensure that the forget gate is activated with a high value. Without specialized training techniques, this can easily lead to overflow and instability in neural networks. To stabilize the networks, we typically need to restrict gradient changes to prevent gradient explosion, often through methods like gradient clipping \cite{arjovsky2016unitary, jing2017tunable}. However, by limiting gradient changes within certain training epochs, we inadvertently place an invisible bound on the forget gate's value. Consequently, neural network may still exhibit short-term memory behavior in practice. To address the problem of short memory in general RNNs for TSF, additional techniques are often necessary. In this paper, we propose to utilize the patching method to overcome this problem as follows.


\subsection{Overall Structure of P-sLSTM}


Figure~\ref{Figure 2: Overall Structure} illustrates the overall structure of the designed P-sLSTM, where multivariate time series data is divided into different channels that share the same backbone but are processed independently. Each channel's univariate series is segmented into patches, processed by a linear layer, and after several system blocks, another linear layer produces the final prediction. Specifically, the batch (\( B \)) of samples \( x \in \mathbb{R}^{L \times M} \) with size \( B \times L \times M \) is initially transformed to \( B \times M \times L \) and undergoes channel independence, resulting in \( (B \cdot M) \times L \). This data is then passed through a patching operator, producing \( (B \cdot M) \times N \times P \), where \( N \) is the number of patches and \( P \) is the patch size. It is projected to \( (B \cdot M) \times N \times \text{Embedding} \) by a linear layer and processed by the xLSTM block. After flattening, the data becomes \( (B \cdot M) \times (N \cdot \text{Embedding}) \) and is finally projected to \( (B \cdot M) \times T \) by a linear layer, with the output reshaped to \( B \times T \times M \).

\subsubsection{Patching}
As we discussed before, we show that although the activation function of sLSTM is adjusted, the model could not have a long memory all the time; hence, it still cannot capture long-term dependencies very successfully. Therefore, inspired by the success of patches in transformer architectures \cite{nie2022time}, we use patches to artificially divide original time series into components and the sLSTM can extract different shorter information; and finally, we can combine them to get global information by linear layers.

\subsubsection{Channel Independence (CI) }

Channel Independence (CI) has been proved to avoid overfitting problems and improve the computational efficiency in Transformer-based models~\cite{nie2022time, zhou2023one} and MLP-based models in time series analysis~\cite{zeng2023transformers, wang2023st}. To our best knowledge, it is the first time to apply CI in RNNs-based models for TSF.

\begin{table*}[!ht]
    \centering
    \begin{adjustbox}{width=1\textwidth}
    {\fontsize{12}{20}\selectfont
        \begin{tabular}{cc|c|cc|cc|cc|cc|cc|cc|cc|cc|cc|cc}
            \cline{2-23}
            & \multicolumn{2}{c|}{Models} & \multicolumn{2}{c|}{{P-sLSTM}} & \multicolumn{2}{c|}{{sLSTM}} & \multicolumn{2}{c|}{{LSTM}} & \multicolumn{2}{c|}{{iTransformer}}& \multicolumn{2}{c|}{{Fedformer}}& \multicolumn{2}{c|}{{Crossformer}}& \multicolumn{2}{c|}{{RLinear}}& \multicolumn{2}{c|}{{DLinear}} &  \multicolumn{2}{c|}{{TimesNet}}& \multicolumn{2}{c}{{Mamba}}
            \\ 
            \cline{2-23}
            &\multicolumn{2}{c|} {Metric}&MSE&MAE&MSE&MAE&MSE&MAE&MSE&MAE&MSE&MAE&MSE&MAE&MSE&MAE&MSE&MAE&MSE&MAE&MSE&MAE \\
            \cline{2-23}
            &\multirow{4}*{\rotatebox{90}{Weather}}& 96 
            & \textbf{0.149} & \textbf{0.208} & 0.172 & 0.261 & 0.236 & 0.304 & 0.162 & \ul{0.210} & 0.217 & 0.296 & \ul{0.160} & 0.229 & 0.175 & 0.225 & 0.176 & 0.237  & 0.172 & 0.220 & 0.219 & 0.276 \\
            &\multicolumn{1}{c|}{}& 192 
            & \textbf{0.197} & \ul{0.256} & 0.253 & 0.337 & 0.262 & 0.330 & \ul{0.205} & \textbf{0.253} & 0.276 & 0.336 & 0.229 & 0.304 & 0.218 & 0.260 & 0.220 & 0.282  & 0.219 & 0.261 & 0.290 & 0.319 \\
            &\multicolumn{1}{c|}{}& 336
            & \textbf{0.249} & \ul{0.297} & 0.338 & 0.394 & 0.341 & 0.387  & \ul{0.265} & \textbf{0.294}& 0.339 & 0.380 & 0.267 & 0.327 & \ul{0.265} & \textbf{0.294} & 0.265 & 0.319  & 0.280 & 0.306 & 0.321 & 0.340 \\
            &\multicolumn{1}{c|}{}& 720 
            & \textbf{0.320} & \ul{0.350} & 0.426 & 0.443 & 0.456 & 0.457  & 0.329 & \textbf{0.339} & 0.403 & 0.428 & 0.354 & 0.394 & 0.329 & \textbf{0.339} & \ul{0.323} & 0.362 & 0.365 & 0.359 & 0.381 & 0.377 \\
            \cline{2-23}
            &\multirow{4}*{\rotatebox{90}{Electricity}}& 96 
            & \textbf{0.130} & \textbf{0.226} & 0.281 & 0.376 & 0.350 & 0.424 & \ul{0.132} & \ul{0.228} & 0.193 & 0.308 & 0.161 & 0.229 & 0.140 & 0.235 & 0.140 & 0.237 & 0.168 & 0.272 & 0.190 & 0.300 \\
            &\multicolumn{1}{c|}{}& 192 
            & \textbf{0.148} & \textbf{0.243} & 0.280 & 0.376 & 0.354 & 0.429 &  0.154 & 0.249 & 0.201 & 0.315 & 0.229 & 0.304 & 0.154 & \ul{0.248} & \ul{0.153} & 0.249  & 0.184 & 0.289 & 0.201 & 0.309 \\
            &\multicolumn{1}{c|}{}& 336
            & \textbf{0.165} & \textbf{0.262} & 0.290 & 0.384 & 0.357 & 0.431 &  0.170 & 0.265 & 0.214 & 0.329 & \ul{0.169} & 0.270 & 0.171 & \ul{0.264} & \ul{0.169} & 0.267  & 0.198 & 0.300 & 0.211 & 0.319 \\
            &\multicolumn{1}{c|}{}& 720 
            & \ul{0.199} & \ul{0.293} & 0.297 & 0.383 & 0.368 & 0.438  & \textbf{0.195} & \textbf{0.288} & 0.246 & 0.355 & 0.248 & 0.340 & 0.209 & 0.297 & 0.203 & 0.301  & 0.220 & 0.320 & 0.219 & 0.323 \\
            \cline{2-23}
            &\multirow{4}*{\rotatebox{90}{Solar}}& 96 
            & \textbf{0.167} & \textbf{0.232} & 0.210 & 0.271 & 0.212 & 0.293 &  0.217 & 0.248 & 0.287 & 0.383 & \ul{0.176} & \ul{0.236} & 0.238 & 0.268 & 0.221 & 0.292 & 0.238 & 0.295   & 0.206 & 0.253\\
            &\multicolumn{1}{c|}{}& 192 
            & \textbf{0.180} & \textbf{0.241} & 0.242 & 0.295 & 0.218 & 0.301 &  0.216 & 0.264 & 0.278 & 0.364 & \ul{0.212} & \ul{0.259}  & 0.263 & 0.283 & 0.249 & 0.308 & 0.223 & 0.285   & 0.232 & 0.292  \\
            &\multicolumn{1}{c|}{}& 336
            & \textbf{0.190} & \textbf{0.248} & 0.244 & 0.308 & 0.272 & 0.335 & 0.220 & 0.271 & 0.319 & 0.397 & \ul{0.218} & \ul{0.263} &  0.284 & 0.295 &  0.268& 0.324 &0.238 & 0.295  &  0.232 & 0.292  \\
            &\multicolumn{1}{c|}{}& 720 
            & \textbf{0.196} & \textbf{0.249} & 0.223 & 0.295 & 0.283 & 0.337 & \ul{0.221} & \ul{0.274} & 0.319 & 0.402 & 0.232 & 0.285 & 0.282 & 0.294 & 0.2708 & 0.3274 &0.245 & 0.290  & 0.234 & 0.294 \\
            \cline{2-23}
             &\multirow{4}*{\rotatebox{90}{ETTm1}}& 96 
            & \textbf{0.292} & \ul{0.343} & 0.631 & 0.552 & 0.771 & 0.659 &  0.302 & 0.365 & 0.326 & 0.390  & 0.344 & 0.384 & \ul{0.301} & \textbf{0.342} & 0.299 & \ul{0.343} &0.341 & 0.378 & 0.458 &  0.453  \\
            &\multicolumn{1}{c|}{}& 192 
            & \textbf{0.329} &0.369 & 0.702 & 0.598 & 0.820 & 0.685  & 0.344 & 0.382 &  0.365 & 0.415  & 0.352 & 0.387 & \ul{0.335} & \textbf{0.363} & \ul{0.335} & \ul{0.365} &0.462 & 0.429 & 0.504 & 0.484 \\
            &\multicolumn{1}{c|}{}& 336
            & \textbf{0.362} & 0.391 & 0.826 & 0.681 & 0.927 & 0.743 & 0.378 & 0.403 & 0.392 & 0.425  & 0.509 & 0.516 & 0.370 & \textbf{0.383} & \ul{0.369} & \ul{0.386} & 0.433 & 0.436 & 0.591 & 0.555\\
            &\multicolumn{1}{c|}{}& 720 
            & \textbf{0.421} &  0.424 & 0.937 & 0.733 & 0.960 & 0.772 & 0.438 & 0.437 & 0.446 & 0.458 & 0.717 & 0.646 & \ul{0.425} & \textbf{0.414} & \ul{0.425} & \ul{0.421} & 0.454 & 0.451 & 0.587 & 0.553\\
            \cline{2-23}
            &\multirow{4}*{\rotatebox{90}{PEMS03}}& 12 
            & \ul{0.077} & \ul{0.183} & 0.103 & 0.203 & 0.111 & 0.214 & \textbf{0.071} & \textbf{0.174} & 0.126 & 0.251 & 0.090 & 0.203 & 0.126 & 0.236 & 0.122 & 0.243 & 0.085 & 0.192 & 0.116 & 0.232  \\
            &\multicolumn{1}{c|}{}& 24 
            & \ul{0.109} & \ul{0.219} & 0.115 & 0.217 & 0.123 & 0.227 & \textbf{0.093} & \textbf{0.201} & 0.149 & 0.275 & 0.121 & 0.240 & 0.246 & 0.334 & 0.201 & 0.317 & 0.118 & 0.223 & 0.162 & 0.268 \\
            &\multicolumn{1}{c|}{}& 48
            & 0.163 & 0.270 & \ul{0.138} & \ul{0.241} & 0.176 & 0.270 & \textbf{0.125} & \textbf{0.236} & 0.227 & 0.348 & 0.202 & 0.317 & 0.551 & 0.529 & 0.333 & 0.425 & 0.155 & 0.260 & 0.247 & 0.334 \\
            &\multicolumn{1}{c|}{}& 96 
            & 0.209 & 0.309 & \textbf{0.163} & \textbf{0.267} & 0.196 & 0.291 & \ul{0.164} & \ul{0.275} & 0.348 & 0.434 & 0.262 & 0.367 & 1.057 & 0.787 & 0.457 & 0.515 & 0.228 & 0.317 & 0.360 & 0.412 \\
            \cline{2-23}
        \end{tabular}
     }
     \end{adjustbox}
     \caption{Multivariate long-term forecasting errors in terms of MSE and MAE. The lower the better. PEMS03 dataset is with forecasting horizon T $ \in \{12, 24, 48, 96\}$. For the others, T $ \in \{96, 192, 336, 720\}$. The best results are highlighted in \textbf{bold} and the second best results are highlighted with an \underline{underline}.}
     \label{Table 2: Main Long-term Forecasting Results} 
\end{table*}

\section{Experiments}
\label{Ch: Experiment}

\subsection{Dataset, Baselines, Metrics and Settings}
We evaluate the performance of proposed P-sLSTM model on 5 popular multivariate datasets, Weather, Electricity, Solar, ETTm1 and PEMS03. These datasets have been extensively used in the previous literature~\cite{wu2023timesnet,liu2023itransformer}.
As baselines, we select SOTA and representative models in the TSF domain, including the following categories: (i) RNNs: sLSTM (without patch), LSTM; (ii) Transformers: iTransformer~\cite{liu2023itransformer}, FEDformer~\cite{zhou2022fedformer}, Crossformer~\cite{zhang2023crossformer}; (iii) MLPs: RLinear~\cite{li2023revisiting}, DLinear~\cite{zeng2023transformers}; (iv) convolutional neural networks (CNNs): TimesNet~\cite{wu2023timesnet}; (vi) SSMs: Mamba~\cite{gu2023mamba}.  Mean Squared Error (MSE) and Mean Absolute Error (MAE) are used to assess the performance of TSF models.

We follow the same data loading parameters (Ex. train/val/test split ratio) from~\cite{wu2023timesnet,liu2023itransformer}. All models use the same experimental setup with prediction length $T \in \{12, 24, 48, 96\}$ and look-back window $L = 96$ for PEMS dataset, following the setting of~\cite{liu2023itransformer}, and $T \in \{96, 192, 336, 720\}$ and look-back window $L = 336$ for other datasets, following ~\cite{nie2022time}. All experiments are conducted on RTX 4090 GPU.

\subsection{Results and Analysis}

\subsubsection{Main Forecasting Results}

The quantitative results for TSF with various baselines are presented in Table~\ref{Table 2: Main Long-term Forecasting Results}. P-sLSTM demonstrates exceptional performance across multiple datasets and prediction length settings, securing 23 first-place and 10 second-place rankings out of 40 settings. 

Notably, among all experimental results, P-sLSTM outperforms sLSTM by 90\% of them and outperforms LSTM in 95\% of the settings, indicating the enhancement of our method over conventional LSTM methods. We also found that P-sLSTM does not work perfectly in PEMS03 dataset. A potential explanation is that PEMS03 is a very noisy series and P-sLSTM does not include any denoising mechanisms for this situation. However, it still has a better performance than many other SOTA methods.


\begin{figure}[!ht]
\begin{center}
\includegraphics[width =0.65 \linewidth]{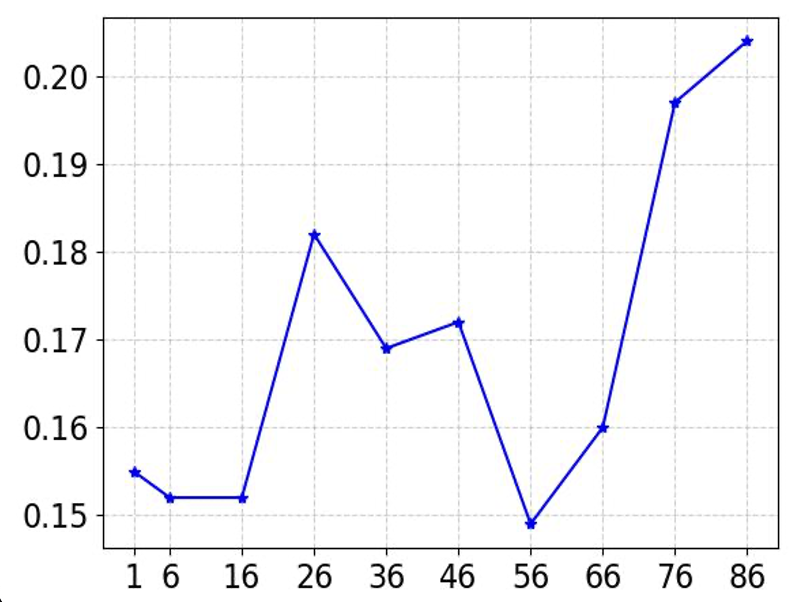}
\end{center}
\caption{Exploration of the patch size on MSE results of P-sLSTM on Weather dataset.}
\label{Patch}
\end{figure}

\subsubsection{Influence of Patch Size}
As you can see from Figure \ref{Patch}, as patch size increases, the prediction accuracy will increase at first, reach an experimental optimal solution, and then the prediction accuracy will decrease as patch size further increases.
Generally, patching will divide the time series sequence into multiple pieces, and each piece contains parts of the sequential information. A small patch will destroy the original temporal sequential information, such that sLSTM cannot process the sequential information. In comparison, a big patch will contain too much information, causing previous information to diminish the model’s ability to effectively integrate new information. This observation aligns with our analysis of the sLSTM's memory properties. However, based on our experiments, for different datasets, there are different optimal patches. Hence, there might be a need to find optimal patches based on trial and error.

\subsubsection{Impact of Look-back Window Sizes}

\begin{figure}[ht]
\begin{center}
\includegraphics[width = 0.6 \linewidth]{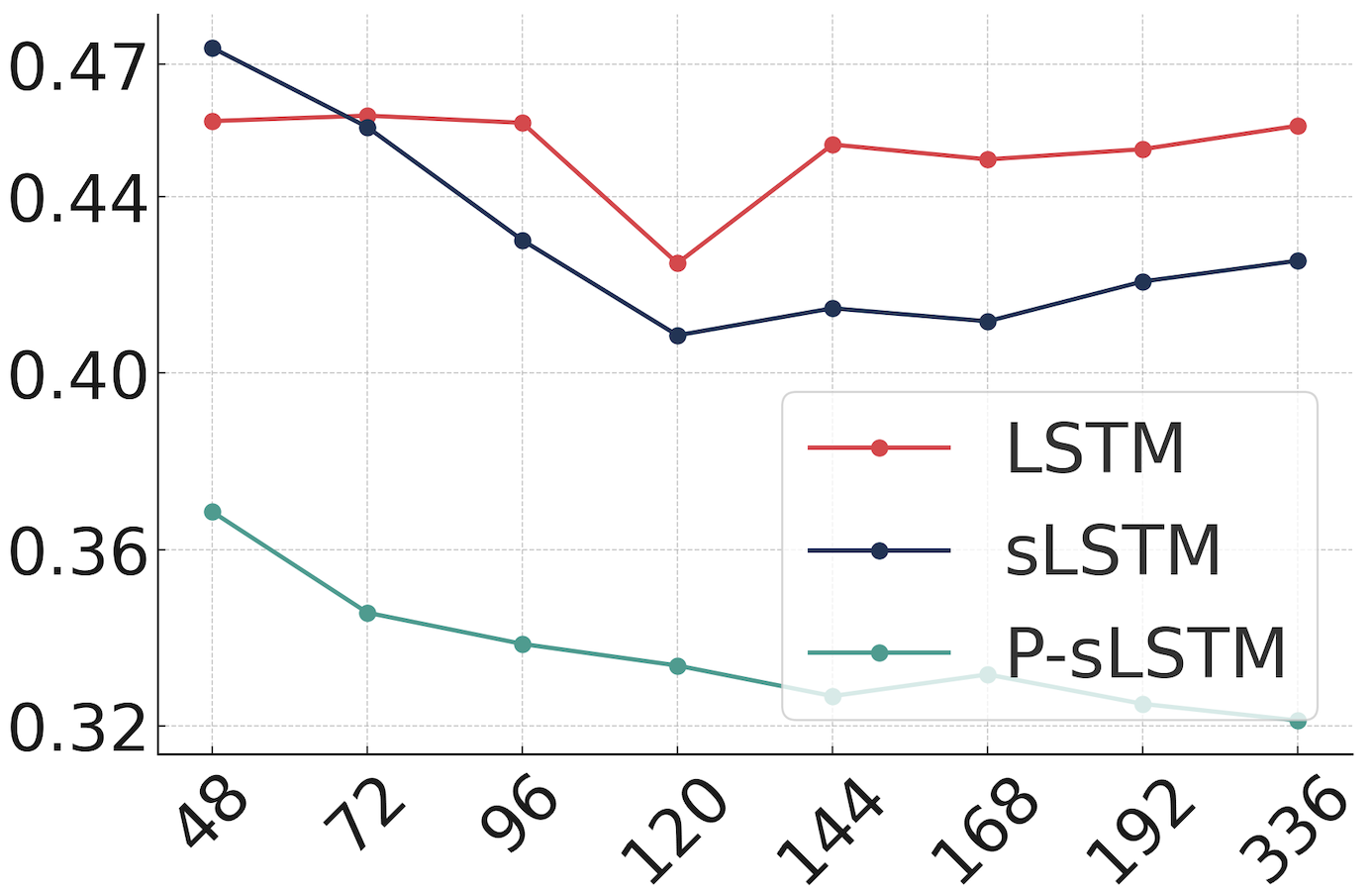}
\end{center}
\caption{The MSE results (Y-axis) of models with different look-back window sizes (X-axis) of long-term forecasting (T=720) on Weather dataset.}
\label{bar}
\end{figure}

As we can see in Figure~\ref{bar}, compared to LSTM and sLSTM, the mechanism of patching can help models capture long-term dependencies. Hence, the prediction accuracy will increase as the look-back window increases.


\begin{table}[!ht]
    \centering
    \setlength\tabcolsep{1mm}
    \small
    {
        \begin{tabular}{cccc}
            \toprule
            Length & Accuracy with M.M. & Accuracy without M.M. \\
            \midrule
            96 & 0.149 / 0.208 & 0.151 / 0.209 \\
            192 & 0.197 / 0.256 & 0.201 / 0.263 \\
            336 & 0.249 / 0.297 & 0.255 / 0.306 \\
            720 & 0.320 / 0.350 & 0.320 / 0.352 \\
            \bottomrule
        \end{tabular}
    }
    \caption{Ablation Study of Memory Mixing (M.M.) on Weather dataset. Values represent MSE/MAE.}
    \label{memory mixing}
\end{table}

\begin{table}[ht]
    \centering
    \setlength\tabcolsep{1 mm}
    \renewcommand{\arraystretch}{0.85}
    \small
            \begin{tabular}{cccccccc}
            \toprule
            \multirow{2}{*}{Dataset} & \multirow{2}{*}{Length} & \multicolumn{3}{c}{CI P-sLSTM} & \multicolumn{3}{c}{CM P-sLSTM} \\
            \cmidrule{3-8}
            & & Train & Val & Test & Train & Val & Test \\
            \midrule
            \multirow{4}{*}{Weather} & 96 & 0.151 & 0.138 & 0.167 & 0.049 & 0.215 & 0.227 \\
            \cmidrule{2-8}
            & 192 & 0.163 & 0.150 & 0.180 & 0.052 & 0.238 & 0.234 \\
            \cmidrule{2-8}
            & 336 & 0.170 & 0.158 & 0.190 & 0.054 & 0.246 & 0.265 \\
            \cmidrule{2-8}
            & 720 & 0.173 & 0.155 & 0.196 & 0.057 & 0.206 & 0.260 \\
            \bottomrule
        \end{tabular}
    \caption{CI vs CM strategies on P-sLSTM.}

    \label{ab2}
\end{table}

\begin{table}[!ht]
    \centering
    \setlength\tabcolsep{0.5mm}
    \renewcommand{\arraystretch}{0.85}
    \small
    {
        \begin{tabular}{lcc}
            \toprule
            Dataset & P-sLSTM (seconds) & iTransformer (seconds) \\
            \midrule
            Weather & 52.54 & 79.12 \\
            ETTm1 & 93.13 & 114.25 \\
            \bottomrule
        \end{tabular}
    }
    \caption{Comparison of Training Time for P-sLSTM and iTransformer across Weather and ETTm1 Datasets.}
    \label{bar1}
\end{table}

\subsubsection{Ablation study of Memory Mixing}
To test the effectiveness of memory mixing of sLSTM in TSF, we conducted an ablation study to remove memory mixing as in normal LSTM. The results of the weather dataset with a horizon of 96 with length 336 are shown in Table~\ref{memory mixing}. As we can see, the memory mixing strategy slightly helps us select important previous temporal information, resulting in an increase in prediction accuracy. However, its improvement is limited. 

\subsubsection{Importance of Channel Independence (CI)}
To test the importance of CI, we prepared a variant of P-sLSTM with channel mixing (CM) and tested its performance. For CI P-sLSTM, the input dimension block of sLSTM is \( (B \cdot M) \times N \times \text{Embedding} \) instead of \( B \times N \times (\text{Embedding} \cdot M) \). The final results using the solar energy dataset are shown in Table~\ref{ab2}. As we can see, CI sLSTM has higher training errors but lower validation errors and testing errors. Hence, we can conclude that CI can significantly prevent overfitting. Therefore, through these experiments, we can conclude that channel independence is not a unique idea in Transformer-based models and MLP-based models. It has the potential to be expanded to more neural network structures. 

\subsubsection{Time Efficiency Study}
We compared the computational efficiency of P-sLSTM with iTransformer (the second best model) for Weather and ETTm1 in Table~\ref{bar1}. P-sLSTM has the lowest computational cost, indicating its potential.



\section{Conclusion}
\label{Ch: conclusion}

In this paper, we develop an LSTM-based method, P-sLSTM, for long term TSF task. We utilize the framework of the sLSTM from NLP in combination with patching to solve the potential short memory problem of original LSTMs or RNNs in general, as well as channel independence techniques to avoid overfitting problems. We anticipate that this work will inspire a renewed exploration of RNN-based/LSTM-based models in TSF tasks, offering valuable insights into RNN structures and applications.

In future work, we might consider more complex patching techniques to preserve as much of the original periodicity of time series as possible. Additionally, there are still known limitations of LSTM/RNNs, such as that they cannot be computed in parallel. To help models perform parallel computing, we can consider adding mLSTM, another LSTM structure that can perform parallel computing.

\section{Acknowledgments}
This work is partially supported by NSFC grant (62250410368), NSFC grant (62402414) and KGR fund (24KKSGR024).

\bibliography{aaai25}

\end{document}